\documentclass[11pt]{article}

\usepackage[]{ACL2023}

\usepackage{times}
\usepackage{latexsym}
\usepackage{graphicx} 
\usepackage[T1]{fontenc}
\usepackage{microtype}

\usepackage{inconsolata}
\usepackage{textcomp}  % Required for encoding \textbigcircle
\usepackage{scalerel}  % Required for emoji \scalerel

\def\green{\scalerel*{\includegraphics{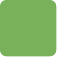}}{\textrm{\textbigcircle}}}
\def\yellow{\scalerel*{\includegraphics{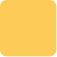}}{\textrm{\textbigcircle}}}
\def\orange{\scalerel*{\includegraphics{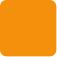}}{\textrm{\textbigcircle}}}
\def\globe{\scalerel*{\includegraphics{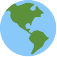}}{\textrm{\textbigcircle}}}
\def\refresh{\scalerel*{\includegraphics{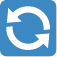}}{\textrm{\textbigcircle}}}
\def\white{\scalerel*{\includegraphics{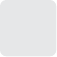}}{\textrm{\textbigcircle}}}

\title{Strategic Insights in Human and Large Language Model Tactics at Word Guessing Games}

\author{Matīss Rikters \\
 Artificial Intelligence Research Center, \\
 National Institute of Advanced \\
 Industrial Science and Technology \\
 \texttt{\{firstname.lastname\}@aist.go.jp} \\\And
 Sanita Reinsone \\
 Institute of Literature, \\
 Folklore and Art, \\
 University of Latvia \\
 \texttt{\{firstname.lastname\}@lulfmi.lv} \\}

\begin{document}
\maketitle
\begin{abstract}
At the beginning of 2022, a simplistic word-guessing game took the world by storm and was further adapted to many languages beyond the original English version. In this paper, we examine the strategies of daily word-guessing game players that have evolved during a period of over two years. A survey gathered from 25\% of frequent players reveals their strategies and motivations for continuing the daily journey. We also explore the capability of several popular open-access large language model systems and open-source models at comprehending and playing the game in two different languages. Results highlight the struggles of certain models to maintain correct guess length and generate repetitions, as well as hallucinations of non-existent words and inflections.
\end{abstract}

\section{Introduction}

An online word puzzle game Wordle, created by Josh Wardle reached its peak popularity in January 2022 \cite {nyt-wordle}. It is a simple daily game where players are challenged to uncover a secret five-letter word using no more than six guesses. After each guess a hint is revealed for each letter, informing the player whether or not the letter is in the correct position, or not in the word at all. Following its widespread popularity, many versions adapted for different languages were developed, including several in the Latvian language. One of them is the focus of our study in this paper.

Wordle's widespread distribution and adaptability to various needs, including educational purposes \cite{Brown2022} and customisation for many languages, offer a valuable opportunity to examine human strategies in the word guessing game alongside those of large language models (LLMs) giving insights into the capabilities and limitations of these models. By analysing how humans play the game, we can identify common patterns and strategies that contribute to successful word guessing. Conversely, evaluating LLM performance on the same task can reveal how well these models mimic human-like reasoning and adaptation.

For human players, there are two main strategies in playing Wordle or similar games. One is to guess word by word and try using the uncovered hints in each subsequent guess by guessing words that contain the letters revealed in hints in those specific positions. The other one is more based on information theory where in the first two or three guesses the player tries to reveal as many hints as possible and then easily guesses correctly in the third or fourth attempt. An example of this in English would be guessing ``OTHER" and ``NAILS", which 10 of the top 12 most frequently used letters in English. In this paper, we take a look at the human side of Wordle-like game playing and analyze how LLMs form their initial guesses and subsequent guesses after hints are provided.

\section{Related Work}

Recent studies \cite{Sweetser2024, Gallotta2024} highlight the significant impact of LLMs on game studies and design. \citet{Sweetser2024} shows the various applications of LLMs, such as improving game AI, development processes, creating dynamic narratives, and understanding player strategies. Similarly, \citet{Gallotta2024} present a detailed survey and road-map for integrating LLMs in games, pointing out their roles in generating intelligent game agents, procedural content, and context-aware narratives. Both studies emphasise the potential of LLMs to revolutionise game design and research by enhancing the understanding of player behaviours and strategies, leading to more immersive and engaging gaming experiences. These reviews collectively suggest that LLMs are well situated to play a major role in the future of game development and interactive storytelling, offering new opportunities for innovation and exploration in the gaming industry.

Examining the computational complexity of games like Wordle more in detail, \citet{Lokshtanov2022} highlight the challenges faced in game design and strategy optimisation. Their study establishes that Wordle is NP-hard, which means it is computationally challenging to determine a winning strategy within a limited number of guesses. This complexity mirrors the intricate problem-solving required for LLMs to effectively handle game instructions. Whereas \citet{chalamalasetti-etal-2023-clembench} assess how chat-optimised large language models cope with following gameplay instructions. Their findings indicate that while current LLMs can follow game instructions to some extent, there is still room for improvement, especially in terms of achieving game objectives efficiently.

Regarding human gameplay, \citet{RiktersReinsone2022BalticHLT} explore the challenges and methodologies involved in adapting a word-guessing game for highly inflectional languages. They find that uncommon inflections, plural word forms, and letters with diacritics/accents or repeating letters make guessing more difficult for players. Meanwhile, common words in nominative singular forms are often the easiest. Whereas \citet{anderson2022} present methods for optimising human strategies in Wordle using maximum correct letter probabilities and reinforcement learning.

\section{Player Survey}

In March 2024, we conducted a voluntary survey to better understand the habits and motivation of the players of one of the Latvian Wordle versions. About 25\% of the daily or weekly players of the game responded with their insights. A non-intrusive link to the survey was placed on the top of the main page for one week, and a total of 110 players noticed it and decided to respond.

The survey contained six questions: play frequency with answer options -- every day, few times per week, few times per month, fewer than once per month); reason for playing -- open-ended; whether it gets easier after playing long term -- yes, no, no opinion; gender; age; and an open text box for any suggestions or comments.

The youngest survey respondent was only four years old, while the oldest was 89. Young people were a minority with only about 12\% up to 20 years old and 27\% between 20 and 40. The majority of players -- 41\% were in the age range of 40 to 60, and the remaining 20\% were over 60. In terms of gender balance, the player base skews more towards females with 64\% vs. 36\% identifying as male. 

The vast majority of around 83\% survey respondents noted that they play every day. A further 10\% said to be playing a few times per week, 2\% -- few times per month, and 5\% -- less often. The answers to this question may, however, be slightly biased towards the more frequent players since the survey was only visible for one week and some potentially less frequent players may not have opened the game specifically in that week.

When it comes to players finding it easier to guess the daily word after having played for a long time -- exactly half said that they don't. Only 35\% said that they find it becomes easier, and 15\% did not have an opinion on the matter.

\begin{figure}
  \centering
  \includegraphics[width=\linewidth]{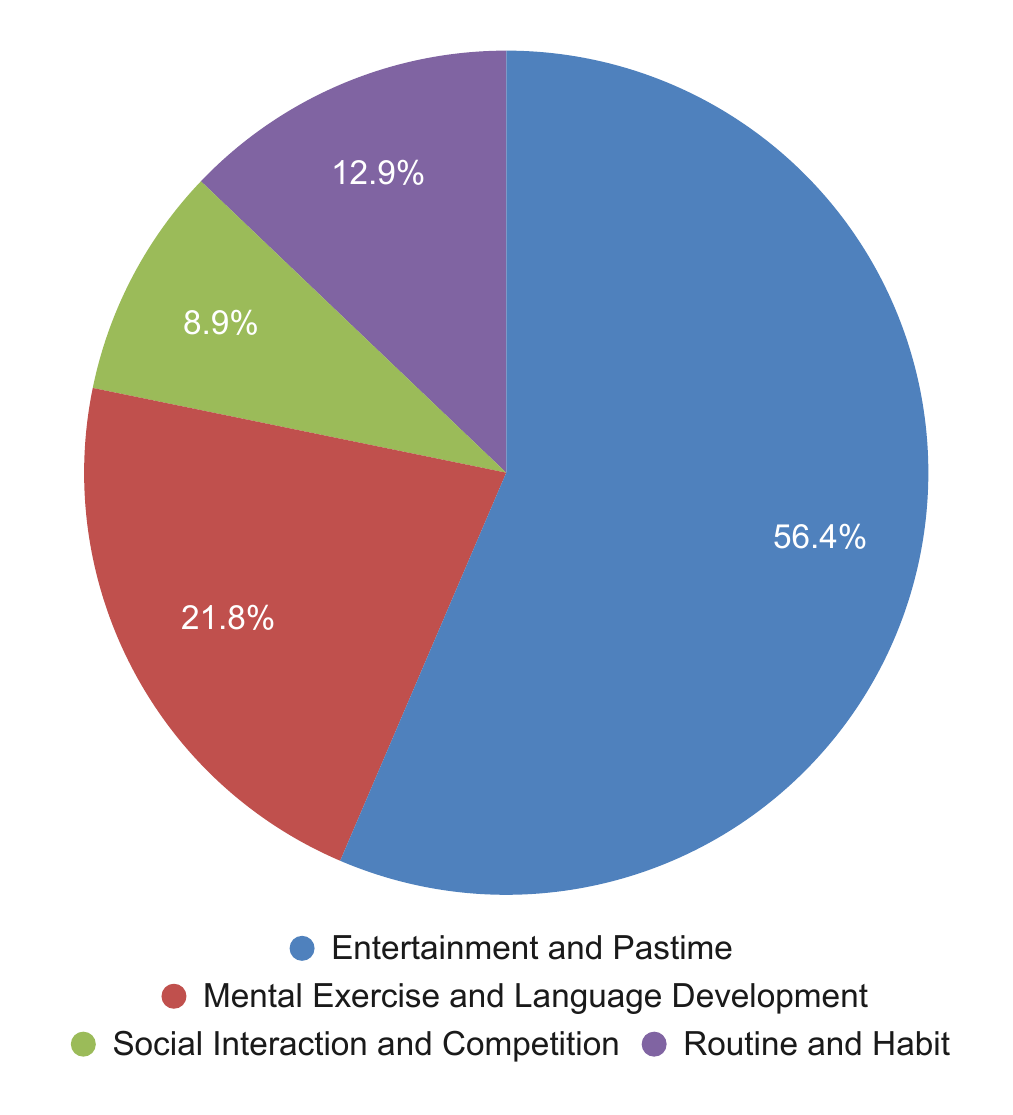}
  \caption{Players' specified reasons for playing the word guessing game regularly.}
  \label{fig:player-reason}
\end{figure}

Figure \ref{fig:player-reason} shows that for the majority of players, the main motivation is `entertainment and pastime'. Interestingly, only 8.9 per cent of respondents specified their motivation as `social interaction and competition', which seemed to be much more important in the first few months when the game was gaining popularity and many players were posting results on Twitter.

\section{Human Gameplay Analysis}

For the purpose of our study, we have acquired a corpus of daily guesses from a Latvian language version of the popular Wordle game provided by its authors. The corpus contains between one and six five-letter words for each play session, the date and time of the session, and the correct answer. With gameplay data collected for over two years, we are able to look into an extended history of trends from human players of the word guessing game.

The statistics show that the most common starter word by far is ``SAULE" (the Sun) used nearly 32,000 times, followed by ``SIENA" (wall) and ``TIESA" (court) at about 14,000 and 13,000 times respectively. When we look at the next words people use after guessing ``SAULE" in the first, it is ``RIPOT" (to roll) at about 1000x, ``TRIKO" at about 700 and ``KORIM" at about 500. Each of these contains five different unique letters which do not overlap with ``SAULE", indicating the utilisation of the information theory-based approach. In fact, the top 6 most frequently used second-guess words after `SAULE" and the top 5 after ``SIENA" all follow this pattern. Overall it seems that roughly 8-10 per cent of players do not overlap letters used between the first two guesses in order to uncover further hints. 

Indeed most players, the authors included, more often than not follow the other strategy of forming each subsequent guess with the hints uncovered in previous guesses. This forces the player to consider the whole list of actual possibilities for the correct answer at each guess, and with a little luck it can actually be more productive than wasting moves to uncover hints for two or three turns and then risking the third or fourth for the ultimate guess. In fact, there is a dedicated ``hard mode" switch in the game which only permits the player to make further guesses that include in the specific positions any letters uncovered in the correct positions by hints.

\begin{figure*}
  \centering
  \includegraphics[width=\linewidth]{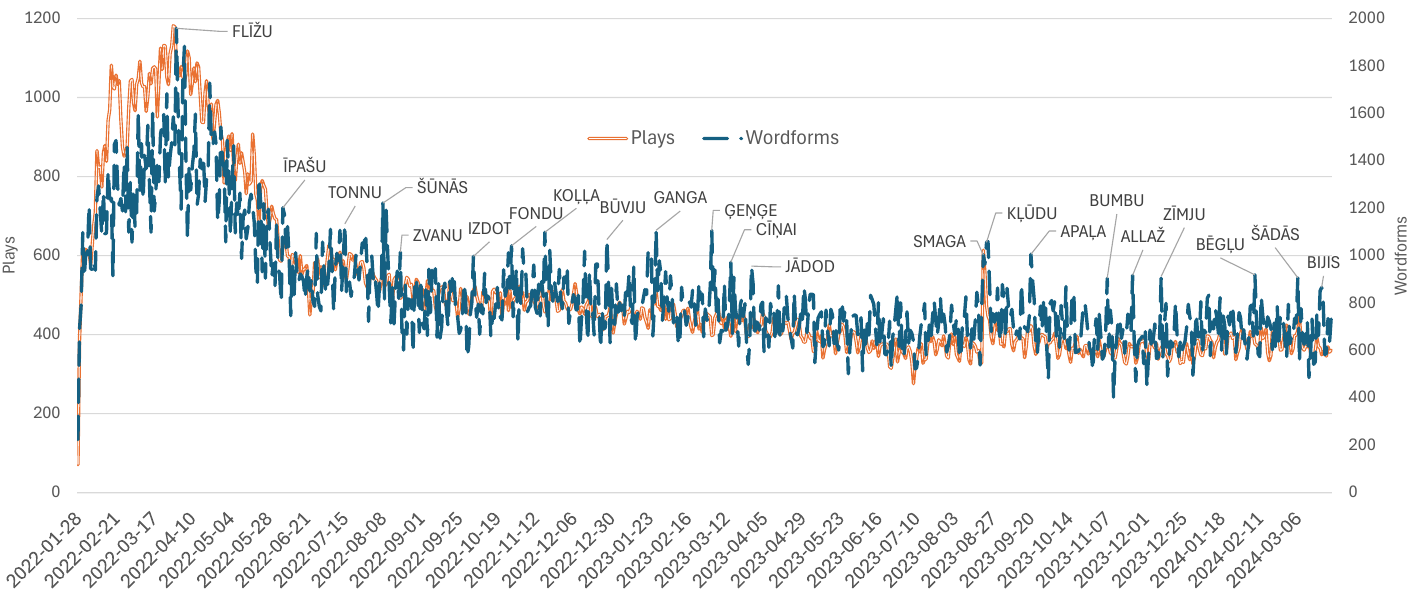}
  \caption{Daily play count and used word form amount changes over the past two years.}
  \label{fig:plays-forms}
\end{figure*}

Figure \ref{fig:plays-forms} highlights how the number of unique game sessions per day stabilised within 6 months after the initial popularity of the game faded. It is contrasted by the number of unique word forms used each day for guessing. The peaks of used word forms show how difficult-to-guess words motivate players to try out less obvious vocabulary and inflections. Most of the correct answers in those peaks are either words with one or more accented letters or old, less commonly used, regional forms of words that the authors included to show up once per week as a special challenge for the players.

\begin{figure*}
  \centering
  \includegraphics[width=\linewidth]{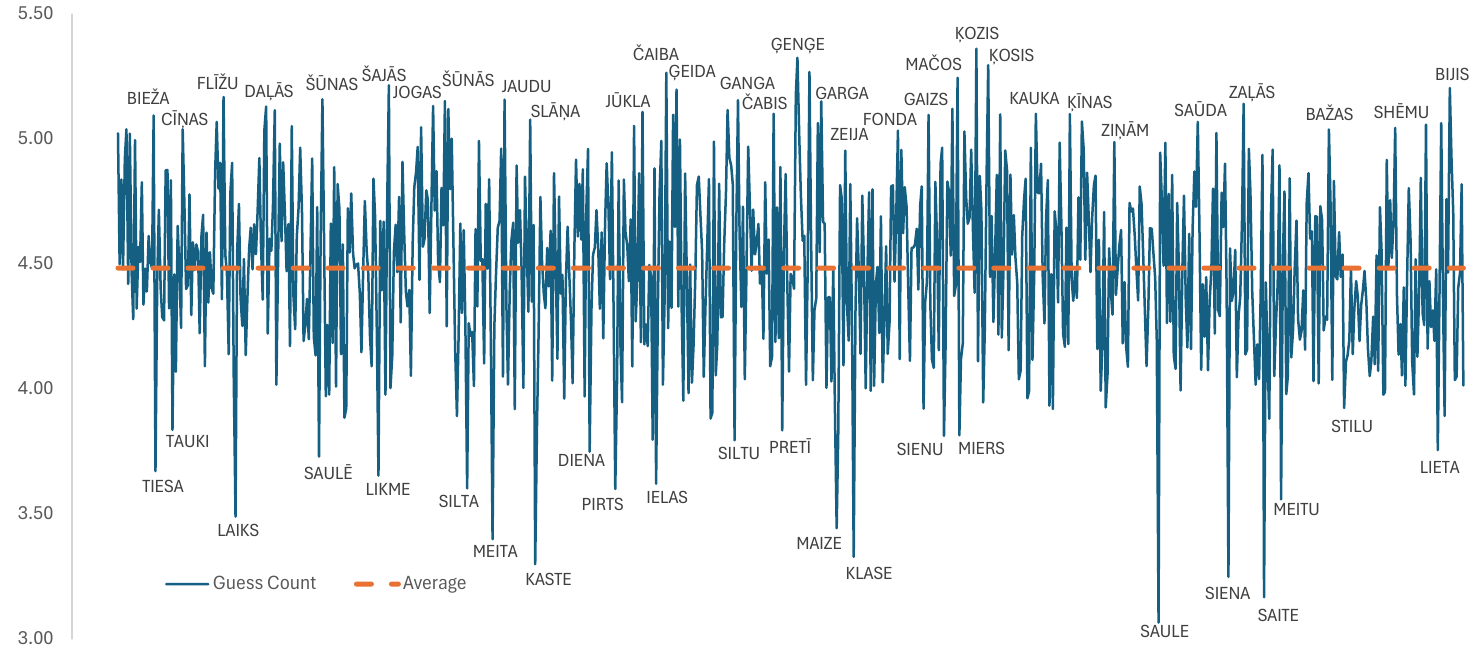}
  \caption{Average numbers of guesses to solve each daily game.}
  \label{fig:average-guesses}
\end{figure*}

Figure \ref{fig:average-guesses} shows the average number of guesses required for solving each daily game. The overall average for the Latvian version of the game is about 4.5, which is somewhat higher than the global average for the English Wordle \cite{WordleStat} at 3.83. The upper peaks in this chart again represent words with accented letters, repeating letters, and old, less commonly used, regional forms of words. The lower peaks are the easiest and are mostly common nouns in nominative singular forms with no repeating letters and no diacritics.

\section{Large Language Model Gameplay Evaluation}

Since modern large language model systems (LLMs) are trained and fine-tuned on a vast amount of web-scraped data, our assumption was that they would be able to guess five-character long words similarly to how humans do it. In our evaluation, we tested five open-access LLM systems: Chat GPT; Gemini and Gemini Advanced; Claude; Mistral; and two open-source models: Llama 3 8B and 70B. We also attempted to use Reka Core\footnote{https://chat.reka.ai/}, however, it was mostly unable to generate five-letter answers and we quickly ran out of available conversation space in the free version.

\subsection{Promting Format}

We first experimented with prompting the LLMs to guess English five-letter words and then followed up by prompting them to guess in Latvian. Table \ref{tab:instruction-table} outlines the instructions we provide to the LLM systems. Initially, we experimented with a slightly different format of the feedback, which would only return five coloured square emoji symbols (for example ``\green \ \yellow \ \green \ \yellow \ \white \ ") to the LLMs (with a proper explanation of each colour in the start instruction prompt), but none of the LLMs were successful in uncovering even the first word this way. Similar to the actual Wordle games, invalid strings which are not actual words in the given language cannot even be entered as potential guesses. Repetitions are permitted, but since they do not uncover further hints, we decided to specifically require the LLMs to generate unique guesses at each turn and not count repeated words towards the six-guess limit.

We were initially planning to experiment with providing both -- some of the most difficult-to-guess words, as well as the easiest words to the LLMs, but after a few preliminary experiments it became clear that they already struggle even with the easy ones. Therefore, we resorted to mainly only experimenting with relatively common words, including only one somewhat difficult word for each language -- ``OVERT" and ``CĪŅAS". The full list of 10 words per each language is shown along with the LLM guess results in Table \ref{tab:llm-result-table}. 

In addition to the regular gameplay script of six subsequent guesses, we offered each LLM one last chance in cases where enough hints were already uncovered during the game that a human player could easily come up with the correct guess. This simulated something like a few-shot prompting approach. For example: ``Guess a five-letter word in English where the second letter is e, the third is a, and the last is h. The word also contains a letter d" or ``Guess the five-letter word: \_I\_AL. One of the two missing letters should be N."

\subsection{Guessing in English}

\paragraph{Claude 3}

We used the publicly accessible Claude 3 Sonnet system\footnote{https://claude.ai/chats - accessed April 2024} for our evaluation. Claude 3 seemed to be the weakest of the commercial LLM systems, only correctly guessing one out of the ten words in our experiments. However, it was the best in the additional task of guessing the word with all uncovered hints provided in a single prompt -- six additional correct guesses. Similar to Gemini Advanced, Claude 3 never repeated previously guessed words in the same turn, nor did it generate non-English words, but it did violate the five-letter length three times.

Among the 10 game sessions, Claude 3 tended to use ``SLATE" as an opening guess four times, and once more as a second guess. Claude mostly generated subsequent guesses seemingly taking previously uncovered hints into account.

\paragraph{Gemini}

For experiments with the Gemini\footnote{https://gemini.google.com - accessed April 2024} systems, we employed both the regular Gemini, as well as the Gemini Advanced (free trial). Overall, Gemini Advanced was the best-performing system with four correct guesses in the first six attempts, two additional correct guesses when prompted with all uncovered hints in a single prompt, and one more relatively close guess. It never attempted to generate non-English words, nor repeat guesses, and only generated one guess which was not five letters long. 

The regular Gemini system was the runner-up also with four correct guesses, albeit mostly different words than the Advanced version, and three relatively close guesses. It did, however, generate the most non-five-letter guesses out of all the systems we compared, and repeated previous guesses 3 times. The regular Gemini system used ``SURGE" as an opening guess twice, while Gemini Advanced used ``TORCH" as a second guess twice. The regular Gemini also used ``DRIFT" as a guess 3x and ``MORAL" 2x. Gemini Advanced seemed to generate subsequent guesses by taking previously uncovered hints into account in most cases, however, the regular Gemini was mostly generating guesses with disregard to the hints. 

\paragraph{GPT 3.5}

We used the publicly accessible GPT 3.5 system within ChatGPT\footnote{https://chatgpt.com - accessed April 2024} for our evaluation. After the two Gemini systems, GPT 3.5 is tied with Mistral for third place with each correctly guessing 2 words within the six attempts, but it was able to guess two more when prompted with all uncovered hints in a single prompt. GPT 3.5 was mostly consistent with keeping guesses at the five-letter limit by only defying it once, but it also generated one non-English word and five repeated words within guesses. 

GPT 3.5 did not use any specific word as a starting guess multiple times, however, it did generate ``RIVER" as the second guess twice, and ``SOLVE", ``SHEET", and ``FAITH" as subsequent guesses twice. GPT 3.5 mostly generated subsequent guesses seemingly taking previously uncovered hints into account.

\paragraph{Llama 3}

We experimented with both sizes of the latest open-sourced\footnote{https://github.com/meta-llama/llama3} version of the Llama models \cite{touvron2023llama} -- Llama 3 8B and 70B. The Llama 3 models seemed weakest overall, not being able to neither fully guess any of the 10 words within the six attempts nor later when prompted with all uncovered hints within a single prompt. They were also the most prone to repetitions of previous guesses, repeating 12 times in the 8B setup and 13 times in 70B. However, the Llama 3 70B model was the only one with all generated guesses being exactly five letters long, as instructed. 

Among the 10 game sessions, both Llama 3 models tended to use ``HOUSE" as an opening guess, with 8B using it a total of 4 times and 70B -- twice. Both Llama models, especially the 8B, struggled to make use of the provided hints at each guess. They kept trying to generate guesses starting with the same two/three letters or ending with the same letter, even if the hints pointed out that those specific letters were not in the word or were in different positions.

\paragraph{Mistral}

We used the publicly accessible Mistral Large system\footnote{https://chat.mistral.ai - accessed April 2024} for our evaluation. Mistral was able to guess ``SKATE" and ``STARE" within the six attempts and was able to guess ``DEATH", ``SAUCE", ``FINAL" and ``TRACE" when prompted with all uncovered hints in a single prompt. Mistral never made any guess that was not an actual English word, it did repeat previously guessed words twice and made four guesses that were not five letters long. 

Among the 10 game sessions, Mistral exhibited a strong preference to use ``TABLE" as an opening guess, using it a total of 7 times and once in a follow-up guess. Mistral seemed to generate subsequent guesses by taking previously uncovered hints into account about half of the time.

\subsection{Guessing in Latvian}

Mistral and the Llama 3 models cannot respond in Latvian when prompted to play the game in Latvian. They did seem to comprehend the instructions and attempt to guess Latvian-sounding words, however, the guesses were almost always non-existent words. Claude 3, Gemini Advanced, and GPT 3.5 were all capable of continuing the conversation and playing the game in Latvian, but they also struggled with frequently providing non-existent words as guesses. 

Out of the 10 Latvian words in our experiment, coincidentally, GPT 3.5 randomly guessed the first one ``SAULE" (the Sun) correctly before any hints were even provided. This is a very common word and a human player's favourite starting guess, so it is not surprising. Meanwhile, Gemini Advanced succeeded in guessing the third word ``LAIKS" (time) after several hints.

\begin{table}
  \centering
  \small
  \begin{tabular}{p{0.2\linewidth} | p{0.7\linewidth}}
  \hline \hline
  \textbf{Meaning} & \textbf{Instruction} \\ \hline \hline
  Start & Let us play a new game. Guess a 5 \\
  instruction & letter word in English. You will have a total of six guesses. Do not repeat a guess. I will reply with a hint for each letter in the guess: correct position; different position; not in the word. Please consider the hints when forming the subsequent guess. Mind that If a guess contains two identical letters if one letter is in the correct position and the hint says that the other is in a different position, it may mean that the answer contains only one of that letter. Answer in one five-letter word. \\ \hline
  Length    & This guess is not 5 letters long. \\ \hline
  Language  & This guess is not a word in Latvian. \\ \hline
  Repetition & This guess was already used. \\ \hline
  Example & d - not in the word \\
  feedback & a - correct position \\
  & r - not in the word \\
  & b - not in the word \\
  & s - correct position \\
   \hline \hline
  \end{tabular}
  \caption{Zero-shot instruction examples provided to the LLM systems.}
  \label{tab:instruction-table}
\end{table}

\begin{table}
% \scriptsize
    \centering
    \begin{tabular}{lcccccc}
         \hline \hline
                        & \green \  & \yellow \    & \orange \  & !5   & \globe \  & \refresh \  \\ \hline
        Gemini Adv.     & 4    & 3           & 1     & 1            & 0           & 0        \\
        Gemini          & 4    & 1           & 3     & 13           & 0           & 3        \\
        GPT-3.5         & 2    & 2           & 0     & 1            & 1           & 5        \\
        Mistral         & 2    & 4           & 1     & 4            & 0           & 2        \\
        Claude-3        & 1    & 6           & 0     & 3            & 0           & 0        \\
        Llama-3 8B      & 0    & 0           & 4     & 7            & 0           & 12       \\
        Llama-3 70B     & 0    & 0           & 2     & 0            & 1           & 13       \\
         \hline \hline
    \end{tabular}
\caption{The results from all English guessing experiments. \green \  represents the number of successful guesses, \yellow \  means that the guesses were quite close and the system/model was able to answer correctly after being presented with all accumulated hints within a single prompt. \orange \  means that given all the same hints, a human could have guessed correctly, but the system/model was unable to. The column !5 shows how many times guesses were made either longer or shorter than 5 letters, \globe \  counts non-English words generated as guesses, and \refresh \  -- repetitions.}
\label{tab:score-table}
\end{table}

\begin{table*}
  % \small
  \centering
  \begin{tabular}{cccccc}
  \hline \hline
  \textbf{Word} & \textbf{Gemini / Advanced} & \textbf{GPT-3.5} & \textbf{Mistral} & \textbf{Claude-3} & \textbf{Llama-3 8B / 70B} \\
  \hline 
  \multicolumn{6}{c}{\textbf{English}}  \\
  \hline 
  SKATE & \green \  / \yellow \  & \yellow \  & \green \  & \yellow \  & \white \  / \white \  \\
  RIDER & \white \  / \green \  & \green \  & \white \  & \green \  & \white \  / \white \  \\
  SAUCE & \yellow \  / \green \  & \white \  & \yellow \  & \yellow \  & \white \  / \white \  \\
  DEATH & \white \  / \yellow \  & \white \  & \yellow \  & \yellow \  & \white \  / \white \  \\
  FINAL & \green \  / \white \  & \white \  & \yellow \  & \white \  & \white \  / \white \  \\
  OVERT & \white \  / \white \  & \white \  & \white \  & \white \  & \white \  / \white \  \\
  TRACE & \green \  / \green \  & \yellow \  & \yellow \  & \yellow \  & \white \  / \white \  \\
  AUDIO & \white \  / \green \  & \white \  & \white \  & \yellow \  & \white \  / \white \  \\
  STARE & \green \  / \yellow \  & \white \  & \green \  & \yellow \  & \white \  / \white \  \\
  PROXY & \white \  / \white \  & \green \  & \white \  & \white \  & \white \  / \white \  \\
  \hline 
  \multicolumn{6}{c}{\textbf{Latvian}}  \\
  \hline 
   & SAULE & TIESA & LAIKS & SIENA & LIETA \\
   & CĪŅAS & BIRZS & REDZE & LAPAS & LAIMA \\
  \hline \hline 
  \end{tabular}
  \caption{Detailed results from the experiments. \green \  represents a successful guess within the six attempts, \yellow \  means that the guesses were quite close and the system/model was able to answer correctly after being presented all accumulated hints within a single prompt, and \white \  indicates an unsuccessful game.}
  \label{tab:llm-result-table}
\end{table*}

\subsection{Result Overview}

None of the systems utilised the guessing approach based on information theory in the first couple of guesses, or even within the first word in some cases. The overall best results were achieved by Gemini Advanced with four main game guesses and three additional ones in the few-shot prompt with all hints provided. On the other hand, both of the open-source Llama 3 models were completely unable to guess a single word.

Out of the 10 relatively common words we selected for our evaluation the most difficult for all LLMs was ``OVERT", followed by ``PROXY", ``AUDIO" and ``FINAL". Meanwhile, ``RIDER", ``SKATE" and ``TRACE" were among the easiest to guess. Among the English words, there were very few cases of generating non-English guesses, but for the Latvian words that was one of the main issues. 

The LLMs rarely repeated guesses, aside from Llama 3 models. In terms of generating answers in the correct length -- most LLMs had few issues, except Llama 3 8B, and especially GPT 3.5.

\section{Conclusion}

In this paper, we presented an analysis of the gameplay strategies of humans and large language models in the context of Wordle games in multiple languages. Our study offers insights into the current capabilities of LLM comprehension and playing abilities of such language-based games. While humans might find such wordplay one of the easiest games in general, we conclude that the LLMs have much room for improvement from this perspective of generalisation. While LLMs such as GPT-3.5, Gemini, and Mistral perform relatively well some of the time, they lack the ability to replicate the nuanced strategies used by human players. More specifically, these models still struggle with generating the correct length words and at times even generate non-existent words, especially when tasked with playing a Wordle game in a language other than English. 

Regarding the human players, the survey data shows that players of Wordle adaptation in Latvian range from young children to seniors, but most fall into the adult 40–60 age group. This diversity of age groups reflects Wordle's broad appeal and its ability to engage players of different ages. Whereas, the high percentage of players who play the game every day shows that the word game has become an integral part of their daily routine. 

In addition, the social and motivational aspects of the game were important to the human players. Many players cited entertainment, mental exercise, and social interaction as the main reasons for getting involved in the game. Players further indicate that they are motivated by the daily competition among family and friends, comparing who can guess the daily word the fastest. This again underlines the multifaceted appeal of Wordle, which combines cognitive challenges with social engagement, making it not only a test of language skills but also a shared cultural experience.

Analysis of human player strategies and habits revealed several interesting patterns. Human players tend to develop specific routines and preferences that improve their performance over time. For example, many players have favourite starting words that they believe increase their chances of finding the correct letters in the fewest possible rounds.

In future work, we plan to explore alternative LLM prompting approaches perhaps enabling more words to be guessed easier. We would also be interested in experimenting with an LLM which is specifically fine-tuned for the Latvian language, but to our knowledge, no such model exists as of writing this paper.

\section*{Limitations}
In this work, we only considered evaluating models and open-access systems that are publicly available at no cost or free trial to help with reproducibility. We plan to prepare a full repository of prompts used in our experiments and publish it on GitHub.

\section*{Ethics Statement}
Our work fully complies with the ACL Code of Ethics\footnote{\url{https://www.aclweb.org/portal/content/acl-code-ethics}}. We use only publicly available models/systems and relatively low compute amounts while conducting our experiments to enable reproducibility. We do not perform any studies on other humans or animals in this research.

\section*{Acknowledgements}
This work has received funding from the project “Research on Modern Latvian Language and Development of Language Technology” (No. VPP-LETONIKA-2021/1-0006).

% \section*{Acknowledgements}

\bibliography{anthology,custom}
\bibliographystyle{acl_natbib}

% \appendix

% \section{Example Appendix}
% \label{sec:appendix}

% This is a section in the appendix.

\end{document}